\documentclass[lettersize,journal]{IEEEtran}

\usepackage[numbers,sort&compress]{natbib}
\usepackage{booktabs}
\usepackage{multirow} 
\usepackage{makecell}
\usepackage{xcolor}
\usepackage{xspace}
\usepackage{colortbl}
\usepackage[table]{xcolor} 
\usepackage{hyperref} 
\hypersetup{
    colorlinks=true,        
    linkcolor=black,        
    citecolor=black,        
    filecolor=black,        
    urlcolor=blue,          
} 

\ifodd 1
\newcommand{\com}[1]{\textbf{\color{red}(COMMENT: #1)}} 
\else

\newcommand{\com}[1]{}
\fi

\def\fig{Fig.}

\def\sysname{\texttt{AirHunt}\xspace}

\def\tab{Table}
\def\eg{e.g.}
\def\ie{i.e.}

\def\eq{Eq.}

\usepackage{amsmath,amsfonts}
\usepackage{algorithmic}
\usepackage{algorithm}
\usepackage{array}
\usepackage{capt-of}
\usepackage[caption=false,font=normalsize,labelfont=sf,textfont=sf]{subfig}
\usepackage{textcomp}
\usepackage{stfloats}
\usepackage{url}
\usepackage{verbatim}
\usepackage{graphicx}
\makeatletter
\let\@oldmaketitle\@maketitle
\renewcommand{\@maketitle}{\@oldmaketitle
    \centering  
    \setlength{\abovecaptionskip}{-3pt}  
    \setcounter{figure}{0}
    \includegraphics[width=1.98\columnwidth]{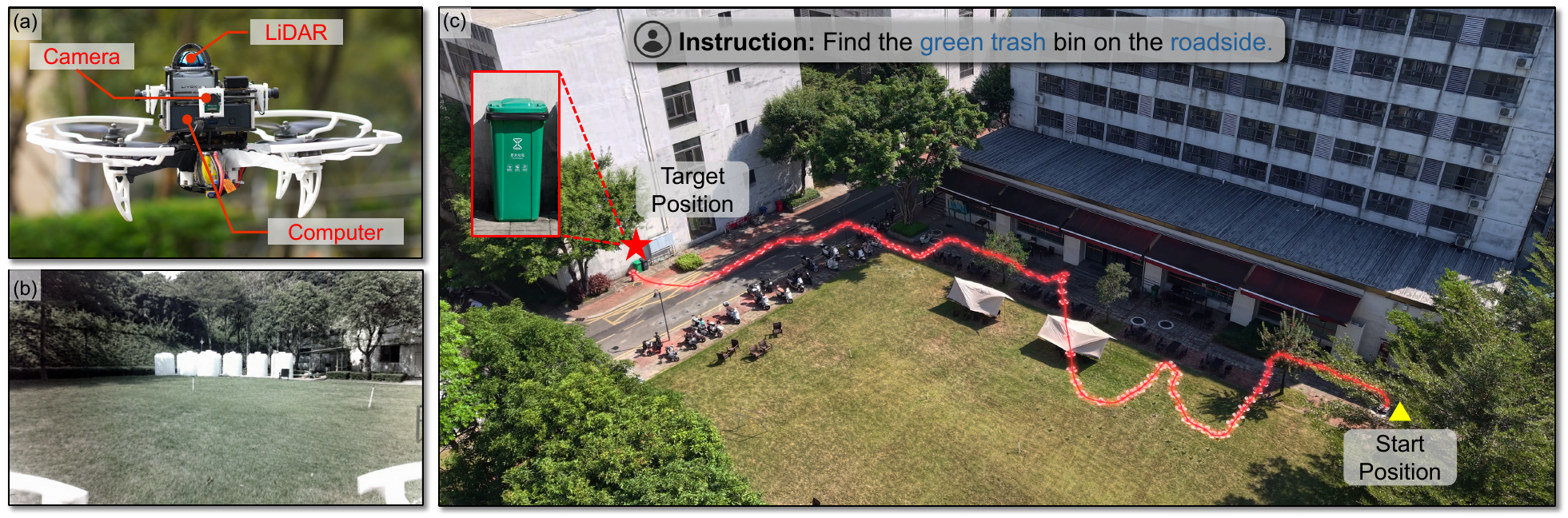}
    \captionof{figure}{
    A drone performs object navigation in a complex outdoor scenario based on a natural language instruction. (a) Custom-built aerial platform equipped with a LiDAR, an onboard computer, and three cameras. (b) Snapshot of the drone's first-person view during navigation. (c) Bird's-eye view of the experimental scenario showing the flight trajectory (red line). The path starts from the corner of the grassy area (yellow triangle) and ends at the target (red star).
    }
    \label{fig:teaser}
    \vspace{-28pt}

}
\makeatother
\title{\fontsize{23pt}{27pt}\selectfont 
AirHunt: Bridging VLM Semantics and Continuous Planning for Efficient Aerial Object Navigation
}

\author{Xuecheng Chen, Zongzhuo Liu, Jianfa Ma, Bang Du, Tiantian Zhang, Xueqian Wang, and Boyu Zhou%
\thanks{Xuecheng Chen, Tiantian Zhang, and Xueqian Wang are with the Shenzhen International Graduate School, Tsinghua University, Shenzhen, China (e-mail: chenxc24@mails.tsinghua.edu.cn; \{zhang.tt, wang.xq\}@sz.tsinghua.edu.cn.)}
\thanks{Zongzhuo Liu and Boyu Zhou are with the Department of Mechanical and Energy Engineering, Southern University of Science and Technology, Shenzhen, China (e-mail: 12432388@mail.sustech.edu.cn; zhouby@sustech.edu.cn).}%
\thanks{Jianfa Ma is with the Department of Computer Science and Engineering, Southern University of Science and Technology, Shenzhen, China (e-mail: 12212225@mail.sustech.edu.cn).}%
\thanks{Bang Du is with the Department of Physics, Southern University of Science and Technology, Shenzhen, China (e-mail: 12312607@mail.sustech.edu.cn).}%
\thanks{Xueqian Wang and Boyu Zhou are co-corresponding authors.}%
}

\begin{document}

\maketitle

\begin{abstract}
Recent advances in large Vision-Language Models (VLMs) have provided rich semantic understanding that empowers drones to search for open-set objects via natural language instructions.
However, prior systems struggle to integrate VLMs into practical aerial systems due to orders-of-magnitude frequency mismatch between VLM inference and real-time planning, as well as VLMs’ limited 3D scene understanding. They also lack a unified mechanism to balance semantic guidance with motion efficiency in large-scale environments.
To address these challenges, we present \sysname, an aerial object navigation system that efficiently locates open-set objects with zero-shot generalization in outdoor environments by seamlessly fusing VLM semantic reasoning with continuous path planning.
\sysname features a dual-pathway asynchronous architecture that establishes a synergistic interface between VLM reasoning and path planning, enabling continuous flight with adaptive semantic guidance that evolves through motion.
Moreover, we propose an active dual-task reasoning module that exploits geometric and semantic redundancy to enable selective VLM querying, and a semantic-geometric coherent planning module that dynamically reconciles semantic priorities and motion efficiency in a unified framework, enabling seamless adaptation to environmental heterogeneity.
We evaluate \sysname across diverse object navigation tasks and environments, demonstrating a higher success rate with lower navigation error and reduced flight time compared to state-of-the-art methods. Real-world experiments further validate \sysname's practical capability in complex and challenging environments. \textit{Code and dataset will be made publicly available before publication.}
\end{abstract}

\vspace{-20pt}
\def\abstractname{Note to Practitioners}
\begin{abstract}
This work addresses a critical challenge: drones struggle to efficiently locate arbitrary objects in large outdoor environments when following natural language instructions (e.g., ``a lost red backpack in the forest''). Existing solutions force drones to hover while waiting for AI (VLM) calculations, disrupting continuous flight and slowing search speed, while over-reliance on these AI models leads to short-sighted decisions, abandoning promising areas too early or revisiting already searched spots. \sysname solves these issues with two key practical improvements: (i) an asynchronous design that lets AI reasoning and flight planning run in parallel, enabling smooth, non-stop flight; (ii) smart logic to use AI only for important camera frames and plan routes that prioritize high-potential areas while avoiding redundant travel. For industry users, this translates to faster search times, higher success rates, and better use of drone battery life, which are critical for real-world missions. Limitations include single-drone operation and network connectivity requirements. Future work could extend it to a multi-robot setup and other autonomous systems.
\end{abstract}
\vspace{-6pt}

\vspace{-15pt}
\section{Introduction}\label{I}
Autonomous target search is a cornerstone of modern drone operations, underpinning critical applications such as disaster response~\cite{schedl2021autonomous}, aerial surveillance~\cite{sun2024moving}, and environmental surveying~\cite{geng2025epic}. 
However, conventional search strategies have been constrained by closed-set detectors and geometry-driven heuristics~\cite{luo2024star,schedl2021autonomous}. The recent emergence of Vision-Language Models (VLMs) has catalyzed a paradigm shift toward open-vocabulary semantic object navigation. This evolution empowers drones to interpret natural language instructions and leverage environmental context, unlocking the potential to fundamentally enhance search efficiency in complex environments.
Despite this promise, integrating VLMs into existing aerial systems remains challenging in practice.

The first challenge stems from the system design of existing VLM-based search systems, which typically treat VLMs as step-wise planners to generate discrete actions~\cite{ji2025towards,xiao2025uav}. Although effective in confined settings, this action-output design introduces two fundamental bottlenecks. 
(\textit{i}) Frequency mismatch: VLMs' high inference latency (often over 2000ms~\cite{wang2025uav}) inherently limits them to sparse action outputs, creating a severe frequency mismatch with drones' real-time planning modules ($>$10Hz, 100ms). Current ``stop-and-infer'' paradigms force drones to hover for VLM responses, disrupting continuous flight, reducing efficiency, and raising mission failure risks in large-scale scenarios given the limited flight endurance.
(\textit{ii}) Limited 3D scene understanding: Since VLMs infer based on 2D visual observations, they often struggle to associate visual entities across viewpoints and integrate them into a global spatial context, resulting in spatially inconsistent actions as the viewpoint shifts. These unstable actions are exacerbated in complex and large-scale outdoor scenarios during long-horizon missions, resulting in premature abandonment of promising regions and redundant revisits.

Secondly, existing planning strategies lack a unified optimization framework that can dynamically balance VLM-derived semantic guidance with geometric motion efficiency. Efficient aerial object navigation requires both effective exploitation of semantic cues and rational consideration of motion costs, yet current methods mostly adopt greedy decisions or simplistic fusion schemes that fail to reconcile both aspects in diverse environments. This absence of systematic balancing induces a binary trade-off in practice: overemphasizing coarse and uncertain semantic guidance results in circuitous routes that exhaust battery resources, while favoring geometric efficiency leads to myopic exploration that overlooks critical regions. This limitation is particularly pronounced in expansive outdoor environments with sparse semantic cues and diverse layouts, severely restricting search performance.

\textbf{Our work}. We tackle the above challenges and propose \sysname, achieving smooth and efficient aerial object navigation in outdoor environments by seamlessly bridging VLM's semantic reasoning with continuous path planning. As shown in \fig\ref{fig:teaser}, \sysname can efficiently locate \textit{open-set} objects with \textit{zero-shot} generalization for various scenarios. \sysname features key designs in both system architecture and algorithms to fully unleash the VLM's potential, as elaborated below.

On the architectural front, we advocate a simple yet effective design philosophy: repositioning the VLM as a high-level dense semantic generator rather than a direct action generator, and integrating semantic information through a 3D representation. 
Since the direct action-generation paradigm with high-latency VLMs cannot meet the high-frequency requirements of real-time flight, we instead use the VLM to predict region-level target-presence probabilities from selected keyframes, producing dense semantic priors that provide more informative guidance for motion planning.
To mitigate limitations in 3D scene understanding, we fuse these language-conditioned priors into a 3D value map as a temporal integrator, which uses confidence-weighted, cross-view integration to reconcile potentially contradictory VLM outputs over time.
Meanwhile, this 3D value map establishes a principled interface that supports a \textit{dual-pathway asynchronous} design between VLM reasoning and path planning, enabling effective synergy while both modules run at their native frequencies and alleviating the latency-mismatch bottleneck.
Specifically, the VLM-driven reasoning pathway asynchronously embeds semantic understanding into the 3D value map, while the high-frequency planning pathway continuously harvests this evolving semantic memory to generate motion trajectories without interruption.

On the algorithm front, we complement the above architecture with two targeted optimizations. 
(\textit{i}) \textit{Active Dual-Task Reasoning (ADTR)}: while the architecture enables asynchronous inference, the continuous drone movement generates high-frequency image streams with redundant content that renders per-frame processing computationally prohibitive. ADTR thus selectively queries the VLM only when observations provide novel geometric coverage or contain instruction-relevant information. This adaptive mechanism turns the VLM into an active reasoner, substantially reducing unnecessary computation while maintaining scene awareness.
(\textit{ii}) \textit{Semantic-Geometric Coherent Planning (SGCP)}: we propose SGCP to resolve the inherent tension between semantic guidance and motion efficiency. SGCP introduces a selective constraint injection mechanism that activates precedence constraints only between regions whose semantic values differ significantly, then minimizes the total travel distance subject to these constraints. This establishes a dynamic partial order: high-value regions are prioritized when their semantic relevance is distinct; when values are comparable, visitation order is flexibly determined by geometric proximity. 
By framing semantic-geometric reconciliation as a unified constrained optimization under value-gap-derived precedence constraints, SGCP adapts seamlessly to environmental heterogeneity, exhibiting semantic-driven behavior in scenarios with distinct high-value regions and geometry-driven efficiency in homogeneous regions.

We conduct extensive experiments in a high-fidelity simulator Unreal Engine~\cite{ue} with diverse object navigation tasks across various environments such as urban downtowns and wilderness villages. 
Evaluation results show that \sysname achieves an average success rate of 73\%, and an average navigation error of 11.6m, outperforming baselines by 49.1\% and 80.3\%, respectively. \sysname also reduces the total flight time by 59.2\%. We further validate our system through over ten hours of real-world experiments, deploying \sysname on a customized quadrotor across diverse and challenging environments, thereby demonstrating its practical effectiveness.

In summary, this paper makes the following contributions.
\begin{itemize}
    \item We propose \sysname, an outdoor open-set aerial object navigation system that repositions VLMs as dense semantic generators rather than direct action generators, and integrates semantics via a 3D value map. It further features a dual-pathway asynchronous design that alleviates latency mismatch and VLMs’ limited 3D scene understanding, enabling continuous flight with adaptive semantic guidance.
    \item We propose ADTR that exploits geometric and semantic redundancy for selective VLM querying; and SGCP, a unified optimization framework that dynamically reconciles semantic priorities and motion efficiency, enabling seamless adaptation to heterogeneous environments.
    \item We evaluate \sysname in a large variety of navigation tasks across various types of environments to demonstrate its zero-shot generalization. We further deploy \sysname in challenging real-world scenarios with extensive evaluation to demonstrate its superior performance. 
\end{itemize}

\vspace{-10pt}
\section{Related Work}

\subsection{Object Goal Navigation}
Object Goal Navigation (ObjectNav)~\cite{batra2020objectnav} is a semantic target-driven task where agents navigate to specified object categories in unknown environments based on egocentric observations.
While early methods require task-specific training (e.g., reinforcement learning~\cite{ye2021auxiliary,chang2020semantic,deitke2022procthor,gireesh2022object}, imitation learning~\cite{ramrakhya2023pirlnav}) for closed-set object categories, recent zero-shot approaches enable generalization to open-set categories by leveraging foundation models in two main ways.
LLM-based approaches use language models for commonsense reasoning. For instance, L3MVN~\cite{yu2023l3mvn} evaluates frontiers by reasoning about object relationships, while ESC~\cite{zhou2023esc} models room-object knowledge as soft logic constraints. 
VLM-based approaches, in contrast, directly extract semantic information from visual observations. For example, VLFM~\cite{yokoyama2024vlfm} constructs language-grounded value maps using cosine similarity scores. More recently, ApexNav~\cite{zhang2025apexnav} adaptively switches between semantic and geometric exploration while maintaining target-centric memory to reduce false detections.
However, existing works predominantly focus on small-scale, semantically homogeneous 2D indoor environments, where structured layouts and semantic hierarchies provide strong cues for object localization. In contrast, our approach extends this task to drone platforms operating in large-scale outdoor scenarios such as urban areas, forests, and waterfront areas. This extension introduces fundamental challenges including extended search ranges, sparse target distribution across unstructured terrain, and the lack of hierarchical semantic organization that indoor methods rely upon. 
\begin{figure*}[t]
    \centering
    \includegraphics[width=0.99\linewidth]{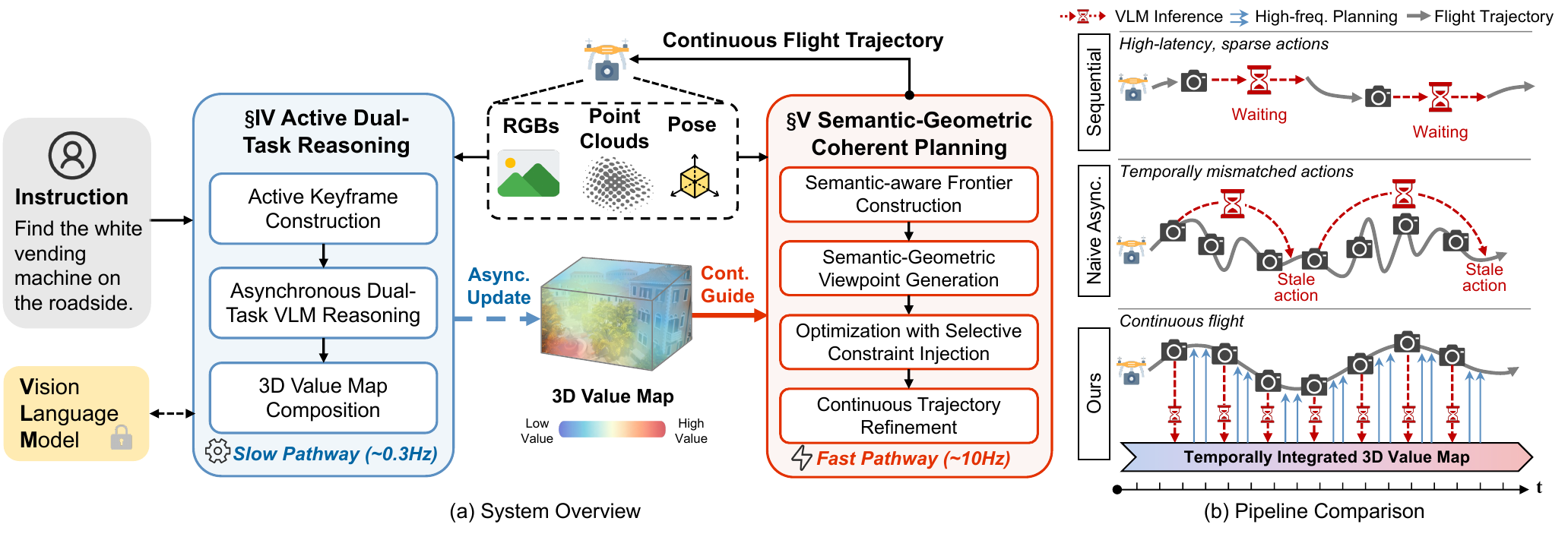}
    \setlength{\abovecaptionskip}{-18pt}  
    \caption{\textbf{System overview and pipeline comparison.}
    (a) \sysname features a dual-pathway asynchronous architecture. The VLM-driven reasoning pathway extracts language-conditioned semantic priors to update a 3D value map asynchronously (Async.), while the high-frequency planning pathway continuously (Cont.) harvests this evolving semantic memory to generate motion trajectories. This design enables both modules to operate at their native frequencies while achieving effective synergy, eliminating mutual blocking.
    (b) Pipeline Comparison. The Sequential approach requires drones to hover for VLM responses, leading to fragmented flight and low search efficiency. The Naive Asynchronous approach enables movement during VLM inference but suffers from action mismatch, as action commands are generated based on stale observations. In contrast, \sysname repositions the VLM as a high-level semantic generator and temporally integrates semantic information into a persistent 3D value map, enabling continuous flight with adaptive semantic guidance.
    }
    \label{fig:sysoverview}
    \vspace{-12pt}
\end{figure*}
\subsection{Aerial Object Search}
In recent years, aerial object search has attracted increasing attention, encompassing both geometric-driven~\cite{luo2024star} and information-driven~\cite{dang2018autonomous} methods. However, these approaches have predominantly relied on closed-set target detectors. While they excel in structured indoor environments, they cannot leverage rich semantic information in scenes to improve search efficiency, nor can they generalize to open-set object categories.
The advancement of embodied intelligence and large language models has introduced new opportunities for drones to understand natural language instructions and locate objects in large-scale, open-world environments. Existing VLM-based methods follow three main paradigms: (\textit{i}) Motion primitive-based selection~\cite{ji2025towards,xiao2025uav}, where VLMs output discrete navigation commands such as moving forward a specified distance based on visual observations; (\textit{ii}) Coordinate selection~\cite{pardyl2025flysearch}, where VLMs identify target locations by choosing from a set of candidate coordinates overlaid on the observed images; and (\textit{iii}) Coordinate prediction~\cite{hu2025see}, where VLMs directly generate 2D pixel coordinates from observed images, which are then projected into 3D space to guide navigation.
While these paradigms demonstrate efficacy in constrained scenarios, they all suffer from stop-and-infer behaviors and over-reliance on VLM decisions, resulting in limited search efficiency when deployed in expansive outdoor environments. 
In contrast, we propose a dual-pathway asynchronous architecture where VLM inference and path planning are decoupled. The planner prioritizes high-value regions inferred by VLMs while minimizing travel distance, generating continuous and consistent flight trajectories that ensure efficient search.

\section{Problem Definition and System Overview}

\subsection{Problem Definition}
This paper addresses aerial object navigation, a variant of the ObjectNav~\cite{batra2020objectnav} task adapted for aerial agents in large-scale outdoor environments. The agent must explore an unknown environment to locate an open-set target object specified by a natural language instruction, while minimizing path length. Task success is achieved when the agent reaches within a predefined distance of the target and executes a stop action.

\subsection{Asynchronous System Architecture}
The system architecture of \sysname is illustrated in \fig\ref{fig:sysoverview}(a). Current practices typically use high-latency VLMs to output step-wise sparse actions, which either induce a stop-and-infer loop or yield temporally mismatched actions, as shown in \fig\ref{fig:sysoverview}(b).
In contrast, \sysname introduces a \textit{Dual-pathway Asynchronous} architecture that establishes a principled interface between VLM reasoning and path planning. 
The core idea is to repurpose the VLM from a direct action generator to a dense semantic signal generator, producing language-conditioned, region-level semantic priors that provide spatially dense guidance.
The VLM and the planner operate at their native frequencies without mutual blocking, with their synergy mediated by a shared intermediate 3D representation (\ie, 3D value map). 
Specifically, the VLM-driven reasoning pathway runs at low frequency to infer these priors from selected observations and asynchronously integrates them into the shared representation. 
Concurrently, the high-frequency planning pathway continuously generates trajectories from the evolving representation, jointly leveraging its semantic priors and geometric information without synchronizing with VLM inference. 
By integrating semantic evidence over time, the shared representation reconciles potential contradictions in VLM outputs and mitigates VLM's limited 3D scene understanding, enabling continuous long-horizon flight with dense semantic guidance rather than fragmented motion driven by sparse, delayed actions.

From the algorithm perspective, \sysname comprises two core modules: the Active Dual-Task Reasoning (ADTR) (\S\ref{IV}) and the Semantic-Geometric Coherent Planning (SGCP) (\S\ref{V}).
ADTR selects informative keyframes and prompts VLM for two tasks: (\textit{i}) infer semantic values across diverse regions to update a 3D value map for exploration guidance; and (\textit{ii}) verify whether detected objects match the task instruction.
Based on the 3D value map, SGCP achieves spatially consistent planning by strategically balancing semantic guidance with geometric travel efficiency.
It first clusters frontiers based on semantic similarity and geometric proximity, then generates viewpoints that prioritize observing high-value regions. Finally, it computes a globally optimal tour that prioritizes semantically valuable regions while minimizing travel distance.

\sysname operates in three stages: (\textit{i}) Initialize: the drone initializes by rotating 360° in place to observe the surroundings; (\textit{ii}) Search: after initialization, the ADTR module continuously updates 3D value maps to guide the SGCP module in generating exploration trajectories; (\textit{iii}) Navigate to target: upon target confirmation by the ADTR module, the system transitions to goal-directed navigation and terminates once the drone reaches sufficient proximity to the target.

\section{Active Dual-Task Reasoning}\label{IV}
VLMs pretrained on large amounts of image-text datasets have internalized extensive commonsense knowledge. However, transferring this semantic understanding to continuous 3D space for navigation remains challenging. This challenge is compounded by the fact that drone-mounted cameras generate high-frequency image streams ($>$20 Hz) with substantial redundancy, making exhaustive per-frame VLM processing computationally prohibitive ($>$2000 ms latency per query).

\begin{figure*}[t]
    \centering
    \setlength{\abovecaptionskip}{-3pt}  
    \includegraphics[width=0.98\linewidth]{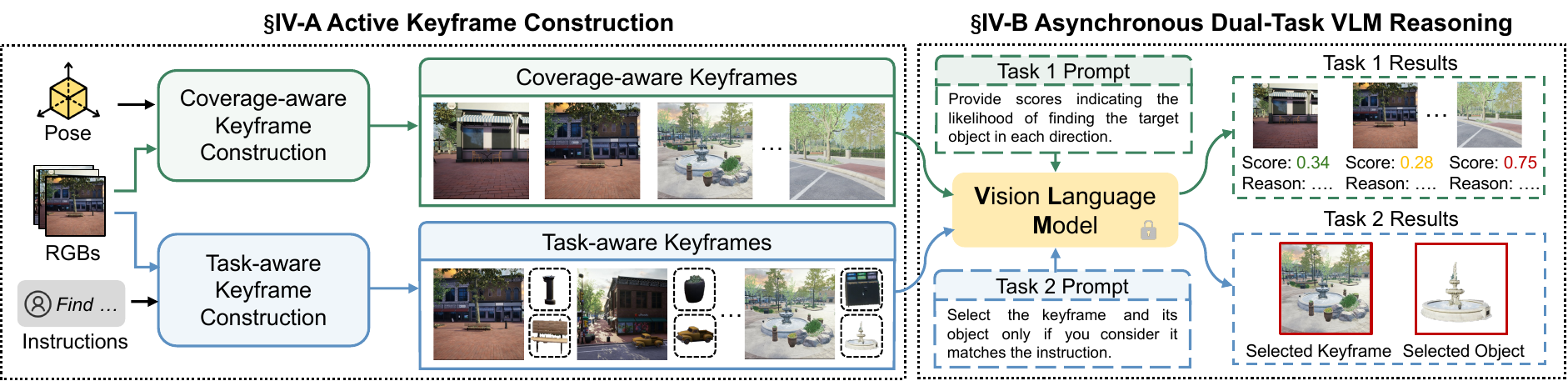}
    \caption{Illustrations of the Active Keyframe Construction and Asynchronous Dual-Task VLM Reasoning Modules. ADTR constructs a coverage-aware keyframe set and a task-aware keyframe set, then queries the VLM for semantic value inference (Task 1) and related object verification (Task 2), respectively.}
    \label{fig:module1}
    \vspace{-10pt}
\end{figure*}

To address these issues, we propose a Active Dual-Task Reasoning (ADTR) module, as illustrated in \fig\ref{fig:module1}. We ground our design on two key observations: 
(\textit{i}) VLMs can directly quantify the semantic relevance of image regions as probabilities conditioned on free-form instructions, providing a quantitative way to identify high-value regions in the drone's planning space; (\textit{ii}) Task-relevant semantic information (i.e., critical objects and cues) in outdoor scenes exhibits high sparsity, which allows for substantial filtering of redundant frames while preserving critical information. We describe the details in the following sections.

\subsection{Active Keyframe Construction}\label{IVA}
To address the computational burden of processing high-frequency image streams from the drone-mounted camera, we adaptively select observed images to construct two types of keyframes: coverage-aware keyframes to maximize geometric coverage of the exploration space, and task-aware keyframes to capture fine-grained and task-relevant information.

\subsubsection{Coverage-aware Keyframe Construction}
We first introduce coverage-aware keyframes to ensure comprehensive geometric coverage of the exploration space. To achieve this, we quantify spatial coverage by discretizing the environment into a 3D voxel grid with resolution $r$. Consider the sequential observations $\mathcal{I} = \{I_1, I_2, \ldots, I_N\}$ from the drone's onboard camera, where each observation $I_i$ consists of an RGB image and its corresponding camera pose. A coverage-aware keyframe is then defined as $\mathcal{K}_s^i = \langle I_i, \mathcal{V}(I_i) \rangle$, where $I_i$ is the observation and $\mathcal{V}(I_i) \subset \mathbb{R}^3$ is the set of 3D voxels visible from the camera pose. The coverage-aware keyframe set at time $t$ is denoted as $\mathcal{S}_t^{\text{cov}}$.

To maximize geometric coverage while minimizing redundancy, we employ a voxel-based overlap criterion. 
Given the current observation $I_t$ and a keyframe $\mathcal{K}_s^i = \langle I_i, \mathcal{V}(I_i) \rangle \in \mathcal{S}_{t-1}^{\text{cov}}$, the overlap ratio between $I_t$ and $I_i$ is defined as $\mathcal{R}(I_t, I_i) = |\mathcal{V}(I_t) \cap \mathcal{V}(I_i)| / |\mathcal{V}(I_t)|$,
where $|\cdot|$ denotes set cardinality. 
For each new observation $I_t$ at time $t$, it is added to the keyframe set if its overlap with all existing keyframes remains below threshold $\tau_{\text{cov}}$, which can be expressed as
\begin{equation}
\mathcal{S}_t^{\text{cov}} = \mathcal{S}_{t-1}^{\text{cov}} \cup \{\langle I_t, \mathcal{V}(I_t) \rangle\}, \ \text{if } \mathcal{R}(I_t, I_i) < \tau_{\text{cov}}, \forall \mathcal{K}_s^i \in \mathcal{S}_{t-1}^{\text{cov}}.
\end{equation}
This criterion ensures that only observations providing sufficient novel geometric coverage are retained, thereby maintaining a compact yet comprehensive representation of the explored space.

\subsubsection{Task-aware Keyframe Construction}
While coverage-aware keyframes ensure comprehensive spatial exploration, they do not guarantee semantic richness relevant to the target object. We therefore introduce task-aware keyframes that are constructed to contain foreground object-level details and contextual scene-level cues pertinent to the instruction, providing the VLM with informative content necessary for accurate target verification.

Specifically, a task-aware keyframe $\mathcal{K}_e^j = \langle I_j, \mathcal{O}(I_j) \rangle$ encodes object-centric scene understanding, where $\mathcal{O}(I_j) = \{o_1, o_2, \ldots, o_M\}$ represents the set of objects detected in observation $I_j$. Each object $o_j$ is characterized by a tuple $(c_j, s_j, \mathbf{x}_j)$, consisting of object category $c_j$, detection confidence score $s_j \in [0,1]$, and 3D position $\mathbf{x}_j \in \mathbb{R}^3$ obtained through object detection, segmentation, and coordinate transformation~\cite{gu2024conceptgraphs}. 
Similarly, we maintain a task-aware keyframe set $\mathcal{S}_t^{\text{task}}$.
Given a natural language instruction $\mathcal{L}$ (\eg, ``find a trash bin on the roadside''), we first query a VLM to identify the top-$K$ object categories $\mathcal{C}_{\text{target}} = \{c_1^*, c_2^*, \ldots, c_K^*\}$ most relevant to the task. For each observation $I_t$, we extract its detected object categories $\mathcal{C}(I_t) = \{c_j \mid o_j \in \mathcal{O}(I_t)\}$. The observation is promoted to a task-aware keyframe if it contains at least one target-relevant object:
\begin{equation}
    \mathcal{S}_t^{\text{task}} = \mathcal{S}_{t-1}^{\text{task}} \cup \{\langle I_t, \mathcal{O}(I_t) \rangle\}, \quad \text{if } \mathcal{C}(I_t) \cap \mathcal{C}_{\text{target}} \neq \emptyset.
\end{equation}
This filtering mechanism discards semantically irrelevant observations while preserving all task-critical visual information for VLM reasoning.

\subsection{Asynchronous Dual-Task VLM Reasoning}\label{IV-B}
The constructed keyframes are fed to our dual-task VLM reasoning module for 
semantic value inference and object verification. To mitigate the variable and potentially long inference latency of VLM queries (sometimes up to 10s under network instability), both tasks execute asynchronously. The VLM reasoning is triggered when either keyframe set reaches its capacity threshold, after which the sets are reset to accumulate new keyframes for the next reasoning cycle.

\subsubsection{Task 1: Semantic Value Inference} 
Due to the high latency of VLM inference, generating per-step action commands from the VLM is impractical for real-time flight. Instead, we use the VLM to reason region-level target-presence probabilities for the scenes captured in keyframes. This probabilistic signal serves as a semantic prior, offering more informative guidance for subsequent motion generation.
Specifically, given the instruction $\mathcal{L}$ and the coverage-aware keyframe set $\mathcal{S}_t^{cov}$, we prompt the VLM to assign a semantic value $v_i \in [0,1]$ to each keyframe $\mathcal{K}_s^i \in \mathcal{S}_t^{cov}$:
\begin{equation}
    v_i = \text{VLM}(\mathcal{K}_s^i, \mathcal{L}),
\end{equation}
where $v_i$ quantifies the probability that the region represented by $\mathcal{K}_s^i$ contains the target object described in $\mathcal{L}$. For instance, given the instruction ``find the white football on the grass'', the VLM assigns higher semantic values to keyframes depicting grassy areas and lower values to keyframes showing basketball courts or buildings. 

\subsubsection{Task 2: Related Object Verification}
Open-vocabulary detectors often produce erroneous labels when objects are occluded, distant, or visually ambiguous. Moreover, they lack contextual understanding to distinguish semantically similar objects (\eg, identifying a black trash bin among bins of different colors). Therefore, we employ the VLM to verify and identify the correct target object from detection candidates based on the given instruction.

Given the task-aware keyframe set $\mathcal{S}_t^{task}$ and instruction $\mathcal{L}$, the VLM selects the target object through:
\begin{equation}
    (\mathcal{K}_e^*, o^*) = \text{VLM}(\mathcal{S}_t^{task}, \mathcal{L}),
\end{equation}
\noindent where $\mathcal{K}_e^*$ denotes the selected keyframe containing the target object, and $o^* \in \mathcal{O}(\mathcal{K}_e^*)$ represents the identified target object from the detection candidates. By integrating both visual appearance cues and spatial context such as object relationships, relative positions, and background information, the VLM performs instruction-specified reasoning to accurately identify the target object.

\subsection{3D Value Map Composition}\label{IV-C}
To guide the drone's search in 3D space, we maintain a 3D value map that encodes the target-presence potential across the search environment based on the semantic values obtained from Task 1. These values are projected into the 3D scene according to the camera's field of view, with occluded regions excluded. A higher value in a region indicates a stronger probability that the target object is located there. By accumulating language-conditioned evidence from multiple viewpoints, the map consolidates potentially inconsistent VLM outputs into a single spatial belief, yielding a more coherent target-probability field for downstream planning and mitigating the VLM’s limited 3D scene understanding.
\subsubsection{Map Representation}
We define a 3D value map $\mathbf{V} \in \mathbb{R}^{W \times H \times D}$ discretized with voxel resolution $r$, comprising two channels: $\mathbf{V} = (\mathbf{V}^{sem}, \mathbf{V}^{conf})$, where $\mathbf{V}^{sem} \in [0,1]$ encodes target-finding potential and $\mathbf{V}^{conf} \in [0,1]$ represents estimation confidence.
For each voxel $\mathbf{v}_k$ at position $\mathbf{x}_k$ visible from drone pose $\mathbf{p}_t$ at distance $d_k = \|\mathbf{x}_k - \mathbf{p}_t\|_2$, the confidence is computed as:
$c_k^t = \max\left(0, 1 - \frac{d_k}{d_{max}}\right)$, where $d_{max}$ is the maximum sensing range. This formulation ensures that confidence decreases with observation distance, reflecting the degradation in visual information quality.

\subsubsection{Map Fusion}
To enable continuous navigation guidance, the value map is updated \textit{asynchronously} as soon as semantic values are returned from Task 1.
Specifically, when the semantic value $v_i$ for keyframe $\mathcal{K}_s^i$ is returned at time $t$, voxels in $\mathcal{V}(I_i)$ are updated via confidence-weighted aggregation~\cite{yokoyama2024vlfm}:
\begin{equation}
    v_k^t = \frac{c_k^{t-1} \cdot v_k^{t-1} + c_k^t \cdot v_i}{c_k^{t-1} + c_k^t},
\end{equation}
where $v_k^{t-1}$ and $c_k^{t-1}$ denote previous values. The confidence is updated to emphasize higher-confidence observations: $c_k^t = \frac{(c_k^{t-1})^2 + (c_k^t)^2}{c_k^{t-1} + c_k^t}$.
This scheme integrates multi-viewpoint information with confidence-based weighting, yielding an effective intermediate representation layer for path planning.

\section{Semantic-Geometric Coherent Planning}\label{V}
Given the 3D value map, planning efficient search trajectories remains significantly challenging due to the inherent conflict between semantic ambiguity and geometric determinism. Unlike dense geometric maps, VLM-derived semantic priors are sparse, probabilistic, and spatially diffuse. Strictly prioritizing high-value regions often triggers oscillatory motion when multiple candidate regions exhibit comparable semantic values, resulting in inefficient circuitous routes. Conversely, prioritizing geometric proximity degrades the system into a blind explorer that delays visiting target-relevant regions. Existing 
approaches lack a principled mechanism to dynamically reconcile these heterogeneous information sources (i.e., semantic and geometric), leading to myopic search behaviors that impair overall search efficiency.

To tackle this challenge, we propose a hierarchical Semantic-Geometric Coherent Planning (SGCP) algorithm that systematically balances semantic guidance with geometric travel efficiency across planning stages. First, we introduce a semantic frontier structure that clusters frontier voxels by both semantic similarity and spatial proximity, enabling fine-grained task-aware exploration beyond purely geometric approaches. Second, we design semantic-geometric viewpoint generation that maximizes information gain weighted by semantic values, ensuring each viewpoint prioritizes target-relevant regions. Third, we develop adaptive constraint-injected global planning that enforces semantic priorities through selective precedence constraints while preserving geometric optimization flexibility, providing flexible search behavior that adapts to diverse environments. This unified design coherently integrates semantic guidance and geometric motion efficiency at frontier, viewpoint, and tour levels, achieving spatially consistent search trajectories with minimal redundant coverage.

\subsection{Semantic-aware Frontier Construction}\label{V-A}
Frontiers are defined as boundary regions between known free space and unknown space~\cite{yamauchi1997frontier}. 
Traditional frontier-based exploration methods leverage this structure to achieve complete scene coverage through purely geometric distance metrics, without considering task relevance~\cite{zhang2024falcon}. 
In contrast, we introduce the semantic frontier that seamlessly integrates geometric information with semantic cues, enabling downstream planners to efficiently explore scenes with fine-grained open-set semantic understanding.

\subsubsection{Semantic Frontier Formulation}
We utilize a volumetric occupancy grid map to represent the environment~\cite{han2019fiesta}, where each frontier voxel $k$ is assigned a semantic value $v_k \in [0,1]$ based on its coordinates in the 3D value map.
A \textit{semantic frontier cluster} is formally defined as $F_i = \langle \mathcal{C}_i, \mathbf{p}_i, s_i \rangle$, where:
\begin{itemize}
    \item $\mathcal{C}_i = \{c_1, c_2, \ldots, c_{n_i}\}$ denotes the set of frontier voxels belonging to cluster $i$,
    \item $\mathbf{p}_i \in \mathbb{R}^3$ represents the geometric centroid of $\mathcal{C}_i$,
    \item $s_i = \frac{1}{|\mathcal{C}_i|}\sum_{c_j \in \mathcal{C}_i} v(c_j)$ indicates the average semantic value of the cluster $i$ and $v(c_j)$ denotes the semantic value of voxel $c_j$.
\end{itemize}

\subsubsection{Incremental Frontier Update}
Each time the occupancy grid map is updated by new sensor measurements, we perform an incremental update of the frontier structure to maintain computational efficiency. We first remove frontier clusters that are affected by the newly observed regions~\cite{zhou2021fuel}. To identify new frontiers, we design a semantic-constrained region growing algorithm.

Specifically, we initiate the region growing process from seed voxels $c_s$ located at the interface between known free space and unknown regions. The algorithm performs breadth-first expansion to aggregate spatially connected frontier cells into clusters~\cite{mehnert1997improved}. Our key innovation is the incorporation of semantic constraints during this process.
Formally, let $\mathcal{N}(c)$ denote the set of spatially adjacent voxels to voxel $c$ in the 6-connected grid. During the expansion of cluster $F_i$ with seed voxel $c_s$, a neighboring voxel $c_j \in \mathcal{N}(c_k)$, where $c_k \in \mathcal{C}_i$, is added to the cluster if it satisfies:
\begin{equation}
    |v(c_j) - v(c_s)| < \tau_s \quad \text{and} \quad c_j \in \mathcal{U},
\end{equation}
where $\tau_s$ is a predefined semantic similarity threshold and $\mathcal{U}$ denotes the set of unvisited frontier voxels. 
Excessively large frontier clusters are then recursively split following~\cite{zhou2021fuel}.
This semantic-constrained expansion enables clustering of frontiers based on both semantic similarity and geometric proximity, thereby facilitating task-informed planning strategies.

\begin{figure}[t]
    \centering
    \includegraphics[width=0.95\linewidth]{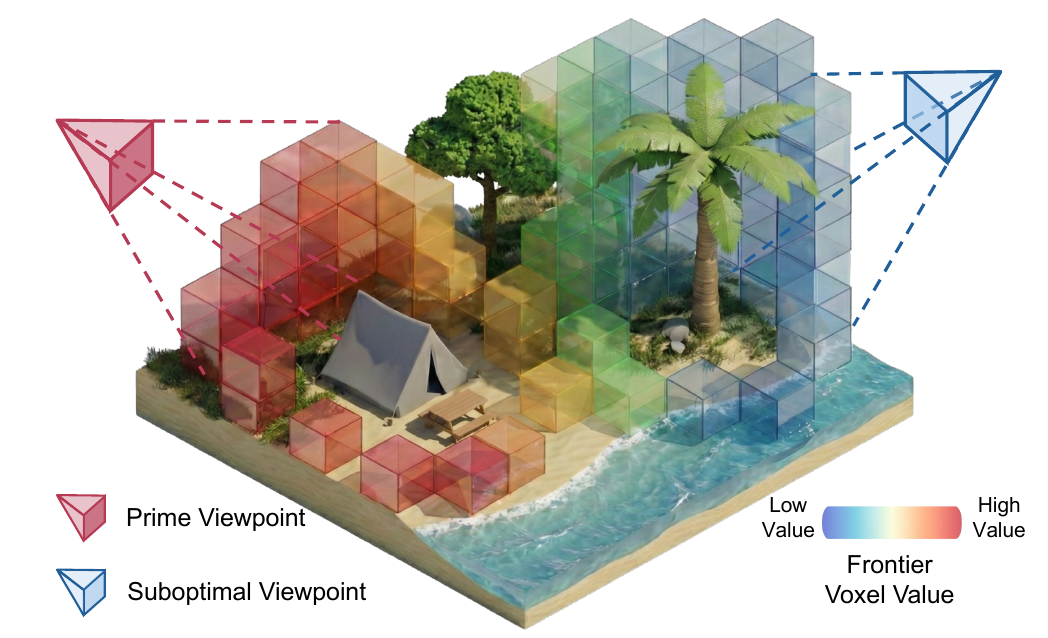}
    \caption{Illustration of Semantic-Geometric Viewpoint Generation. Given the instruction ``Fly to the tent on the beach'', frontier voxels near the target are assigned higher semantic values (red). The \textit{prime viewpoint} is selected to maximize information gain, computed as the sum of semantic values over observable frontier voxels, thereby focusing on the target tent.
 }
    \label{fig:module2-1}
    \vspace{-12pt}
\end{figure}

\subsection{Semantic-Geometric Viewpoint Generation}\label{V-B}
For each frontier cluster $F_i$, we generate a set of candidate viewpoints around the cluster's centroid $\mathbf{p}_i$ to enable observation of the unexplored regions~\cite{zhou2023racer}.
This yields a viewpoint set $\mathcal{U}_i = \{u_i^1, u_i^2, \ldots, u_i^{M_i}\}$, where each viewpoint $u_i^j = (\mathbf{p}_i^j, \theta_i^j)$ consists of position $\mathbf{p}_i^j$ and yaw angle $\theta_i^j$.

To select the most informative viewpoint for each cluster, we evaluate each candidate based on both geometric coverage and semantic importance, as illustrated in \fig\ref{fig:module2-1}. For each candidate viewpoint $u_i^j$, we first evaluate its coverage by identifying all frontier voxels within the sensor's field of view (FoV) that are not occluded by occupied voxels. We then quantify the information gain as the sum of semantic values over all observable voxels as $S(u_i^j) = \sum_{c_k \in \mathcal{N}_{obs}(u_i^j)} v(c_k),$
where $\mathcal{N}_{obs}(u_i^j)$ denotes the set of observable frontier voxels from viewpoint $u_i^j$.
The viewpoint with the highest information gain is selected as the \textit{prime viewpoint} $u_i^*$ for cluster $F_i$.
This approach ensures that the prime viewpoint not only maximizes the number of observable frontier voxels but also gives priority to voxels with higher semantic values, facilitating exploration of informative areas.

\begin{figure}[!t]
    \centering
    \setlength{\abovecaptionskip}{-3pt}  
    \includegraphics[width=0.98\linewidth]{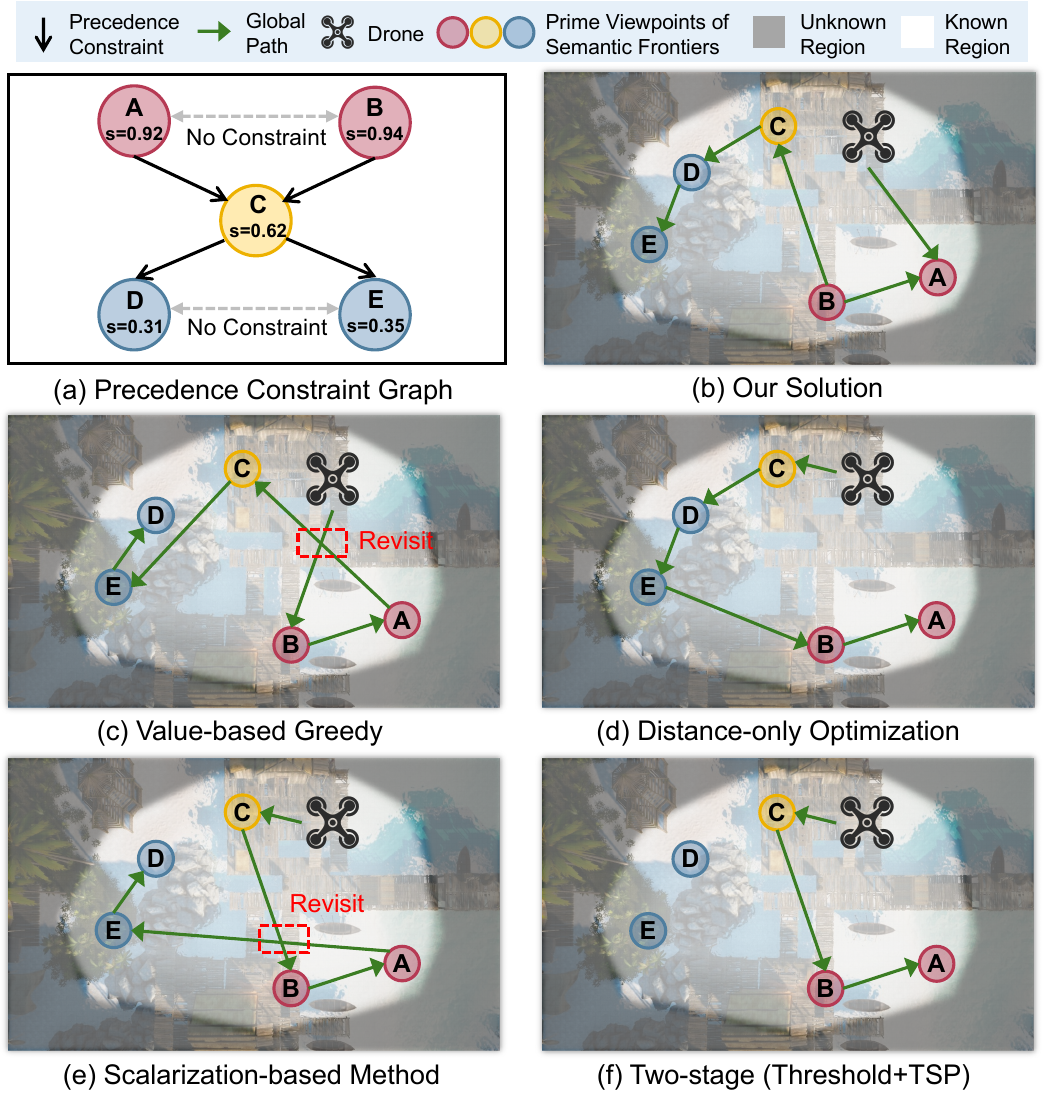}
    \caption{
    Comparison of global planning strategies. (a) The constraint graph selectively enforces precedence only when semantic value differences are significant (e.g., A(0.92)$\rightarrow$C(0.62)), allowing geometric optimization for similar-value frontiers (e.g., A and B). (b) Our solution generates a globally coherent path that prioritizes high-value frontiers without revisits. (c)-(f) Baseline approaches tend to yield revisits of previously searched regions (red dashed boxes) or delay visits to semantically critical regions (A and B).
    }
    \label{fig:module2-2}
    \vspace{-12pt}
\end{figure}

\subsection{Optimization with Selective Constraint Injection}\label{V-C}
Given a set of semantic frontier clusters, our objective is to compute a global exploration tour that prioritizes target-relevant regions while minimizing total traversal time. Existing planners typically optimize one side or combine both via heuristics, but neither yields consistently efficient tours. Value-based greedy selection (\fig\ref{fig:module2-2}(c)) ignores spatial structure and induces large detours when high-value frontiers are dispersed, whereas distance-only optimization (\fig\ref{fig:module2-2}(d)) prioritizes short moves and can defer semantically critical regions. Hybrid heuristics mainly adopt two paradigms. (\textit{i}) scalarization method greedily maximizes a utility score (e.g., value/distance), which tends to bias toward nearby moderate-value frontiers and postpone distant yet substantially higher-value regions (\fig\ref{fig:module2-2}(e)); (\textit{ii}) two-stage method thresholds semantic values to form a frontier subset and then solves a distance-optimal tour. This hard boundary may discard useful regions, and the distance-driven ordering within the subset does not preserve semantic priorities, providing no guarantee that higher-value regions are visited earlier (\fig\ref{fig:module2-2}(f)).

To address this challenge, our key idea is to \textit{selectively} inject ordering constraints only between frontier clusters whose semantic values differ significantly (\fig\ref{fig:module2-2}(a)), and the frontier visitation order is then optimized by minimizing the geometric traversal cost subject to these constraints. 
This design adapts naturally to the semantic value distribution: when a region is distinctly more target-relevant, dense precedence relations enforce early visitation; when semantic scores are comparable, the constraint set becomes sparse, allowing the visiting order to be dominated by geometric proximity. 
By reconciling semantics and geometry through a unified optimization framework with value-gap-derived precedence constraints, SGCP achieves semantic-driven exploration in scenarios with distinct high-value regions and geometry-efficient routing in semantically homogeneous ones, without requiring environment-specific parameter tuning.
This formulation can be transformed into a Sequential Ordering Problem (SOP)~\cite{hernadvolgyi2004solving}, which can be efficiently solved by off-the-shelf solvers. We provide the formal definition of the SOP and present our formulation and solution below.

\textit{Definition 1 (Sequential Ordering Problem)}:
Given a directed graph $G = (V, E)$, edge weight $w : E \to \mathbb{R}$ and a set of precedence constraints $\mathcal{P} \subseteq V \times V$, the SOP searches for a minimum cost permutation of vertices from $s \in V$ to $t \in V$ that satisfies the precedence constraints.

\subsubsection{Selective Precedence Constraint}
In our case, for each cluster $F_i$, we define a vertex $v_i \in V$ locates at its prime viewpoint $u_i^*$. 
We then enforce visiting priorities of high-value frontiers through precedence constraints that adapt to the semantic value distribution.
The precedence constraints set is defined based on semantic value differences:
\begin{equation}
    \mathcal{P} = \{(v_i,v_j) \mid s_i - s_j > \rho \},
\end{equation}
where $s_i$ is the semantic value of cluster $F_i$, and $\rho$ is a positive threshold controlling sensitivity to semantic differences. 
The precedence constraint $(v_i,v_j) \in \mathcal{P}$ enforces that the prime viewpoint $u_i^*$ of cluster $F_i$ (with higher semantic value) must be visited before the prime viewpoint $u_j^*$ of cluster $F_j$.
This design ensures that only semantically significant differences trigger ordering constraints (\fig\ref{fig:module2-2}(a)), allowing geometric optimization to dominate when semantic cues are weak.

\subsubsection{Geometric Cost Calculation}
While precedence constraints handle semantic priorities, geometric costs determine the actual traversal efficiency when semantic cues are weak.
For any two viewpoints $u_i = (\mathbf{p}_i, \theta_i)$ and $u_j = (\mathbf{p}_j, \theta_j)$, we define the geometric cost (\ie, edge weight $w$) as:
\begin{equation}
    C(u_i, u_j) = \max\left\{\frac{d(\mathbf{p}_i, \mathbf{p}_j)}{v_{\max}}, \frac{|\theta_i - \theta_j|}{\omega_{\max}}\right\},
\label{eq:cost}
\end{equation}
where $d(\mathbf{p}_i, \mathbf{p}_j)$ is the path length computed using A$^*$ algorithm on the occupancy grid map, and $v_{\max}$, $\omega_{\max}$ are the maximum translational and angular velocities, respectively. The geometric cost between two frontier clusters is then defined as the cost between their prime viewpoints:
\begin{equation}
    C_{\text{geo}}(F_i, F_j) = C(u_i^*, u_j^*),
\end{equation}
where $u_i^*, u_j^*$ are the prime viewpoints of clusters $F_i, F_j$.

\subsubsection{Cost Matrix Integration}
We construct a cost matrix $\mathbf{C}_G \in \mathbb{R}^{(N_c+1) \times (N_c+1)}$ that integrates geometric costs and semantic constraints:
\begin{equation}
\mathbf{C}_G = \begin{bmatrix}
c_{c,c} & \mathbf{c}_{c,k}^T \\
\mathbf{c}_{k,c} & \mathbf{C}_{k,k}
\end{bmatrix},
\end{equation}
where subscript $c$ denotes the current position and $k$ denotes frontier clusters.

The core component $\mathbf{C}_{k,k} \in \mathbb{R}^{N_c \times N_c}$, which encodes costs between frontier clusters, is defined as:
\begin{equation}
    \mathbf{C}_{k,k}(i,j) = 
    \begin{cases}
    -1 & \text{if } (i,j) \in \mathcal{P} \text{ and } i\ne j, \\
    0 & \text{if } i = j,\\
    C_{\text{geo}}(F_i, F_j) & \text{otherwise}. \\
    \end{cases}
\end{equation}
This unified cost matrix allows the solver to adaptively respect semantic priorities (via precedence set $\mathcal{P}$) while minimizing geometric travel costs where no constraints apply.

The vector $\mathbf{c}_{c,k} \in \mathbb{R}^{N_c}$ represents costs from the current position to each frontier cluster, computed using \eq\ref{eq:cost}.
To ensure the tour starts from the current position, we set $\mathbf{c}_{k,c} = -\mathbf{1}_{N_c}$ and $c_{c,c} = 0$, creating precedence constraints that require all frontier clusters to be visited after the starting position. 

\begin{figure*}[t]
    \centering
    \setlength{\abovecaptionskip}{-7pt}  
    
    \includegraphics[width=0.99\linewidth]{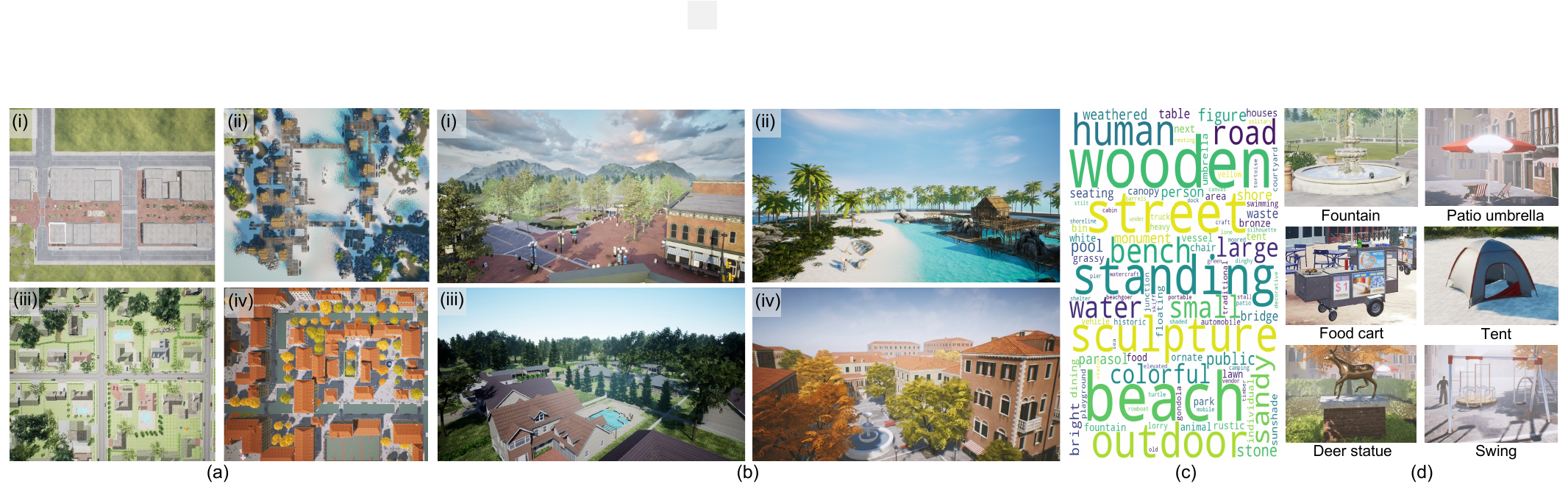}
    \caption{(a) Top-down views and (b) snapshots of four simulation environments: (i) Downtown, (ii) Cabinlake, (iii) Neighborhood, and (iv) Venice. (c) Word cloud of task instructions. (d) Images of example target objects.}
    \label{fig:simenv}
    \vspace{-12pt}
\end{figure*}

\subsection{Continuous Trajectory Refinement}\label{V-D}
During the exploration phase, once the global cluster visitation order is 
determined, we refine the local viewpoint selection across consecutive frontier clusters to minimize trajectory execution time~\cite{zhou2021fuel}. 
Upon target confirmation from the ADTR module, the system transitions to goal-directed navigation, generating a viewpoint around the verified target position with direct line-of-sight. 
Given the next viewpoint to be visited, whether in exploration or goal-directed navigation, we generate smooth, continuous B-spline trajectories that optimize both smoothness and execution time while satisfying safety and dynamic feasibility constraints~\cite{zhou2019robust}. 
The complete trajectory generation process runs in real-time, with replanning triggered whenever new frontiers are discovered or the 3D value map is updated, enabling efficient exploration in partially unknown outdoor environments.

\section{Benchmark and Analysis}

\subsection{Experimental Setup}
\noindent \textbf{Simulation Environments}.
We evaluate our system in outdoor scenarios using the high-fidelity simulator Unreal Engine 4.27~\cite{ue}. The benchmark comprises four diverse large-scale environments: Downtown, Cabinlake, Neighborhood, and Venice. These environments contain various terrain types including villages, cities, parks, forests, roads, seas, and mountains, covering areas from 160,000 to 562,500 $\mathrm{m}^2$. 
The benchmark includes 85 task episodes involving 25 target object categories, described by open-ended task instructions that incorporate environmental semantics (e.g., boats on water, cars on roads).
The environments, word cloud of task instructions, and representative objects are shown in \fig\ref{fig:simenv}.

\noindent \textbf{Metrics}.
Following prior works~\cite{xiao2025uav,ji2025towards}, we adopt five common metrics: Success Rate (SR), Oracle Success Rate (OSR), Success weighted by inverse Path Length (SPL), Distance to Goal (DTG), and Flight Time (FT). SR measures whether the drone stops within a threshold distance of the goal object. OSR checks if the drone comes within this threshold at any point during the episode. SPL evaluates navigation efficiency by weighting success with the ratio of optimal to actual path length. DTG computes the final Euclidean distance to the target object. FT records the total flight time per episode from takeoff to task termination, including the VLM inference time. Note that FT is a critical metric for real-world deployment that has been overlooked in previous works.

\noindent \textbf{Baselines}. 
We evaluate \sysname against a diverse set of drone-based object navigation methods. 
\textbf{PRPSearcher}~\cite{ji2025towards} is a motion primitive-based pipeline that maintains a semantic attraction map and leverages VLM reasoning for discrete action selection.
\textbf{UAV-on}~\cite{xiao2025uav} uses a VLM to convert RGB frames into text, augments them with depth and pose history, and passes the structured prompt to an LLM to produce discrete navigation commands. \textbf{FlySearch}~\cite{pardyl2025flysearch} overlays RGB images with a coordinate grid and queries a VLM to select the next target location. \textbf{STARSearcher}~\cite{luo2024star} relies purely on geometric information to guide exploration, using a hierarchical planner to seamlessly switch between exploring unknown areas and inspecting surface regions. 
\textbf{Human Agent} represents the average performance of three professional drone operators who completed over 20 hours of training in the simulator, establishing the human expert baseline.

\noindent \textbf{Implementation Details}. 
We simulate a drone in AirSim~\cite{shah2017airsim} equipped with a LiDAR sensor (10 m range) and three RGB cameras (facing front, left, right), each with a $90^\circ \times 60^\circ$ field of view. Dynamic limits are set to $v_{m}=2.0$ m/s, $\mathbf{a}_{m}=1.5$ m/s$^2$, and $\omega_{m}=0.9$ rad/s. Our system uses Grounding-DINO~\cite{liu2024grounding} for object detection and SAM2~\cite{ravi2024sam2} for segmentation. All methods employ Qwen-VL-Max API~\cite{yang2025qwen3} as the VLM for reasoning to ensure fair comparison. Semantic threshold $\rho$ is set to 0.15. For evaluation metrics, the SR and OSR adopt a distance threshold of $5\text{m}$, while each task is limited to a maximum of $50$ VLM queries. Each task episode runs three times per method on an identical computer with an Intel Core i7-14700KF, GeForce RTX 4070 12G, and 32 GB memory. We employ an open-source solver~\cite{solver} to solve the optimization problem in \S\ref{V-C}.

\begin{table}[t]
\caption{\textbf{Overall Performance Comparison}. Best results in \colorbox{cyan!15}{cyan}, second-best in \colorbox{violet!15}{violet}, and Human Agent in \colorbox{gray!20}{gray}.}
\vspace{-5pt}
\centering
\renewcommand{\arraystretch}{1.3}
\resizebox{\columnwidth}{!}{%
\begin{tabular}{lcccccc}
\toprule
\textbf{Method} & \textbf{SR $\uparrow$} & \textbf{OSR $\uparrow$} & \textbf{SPL $\uparrow$} & \textbf{DTG $\downarrow$} & \textbf{FT $\downarrow$} \\
\midrule
Human Agent &\cellcolor{gray!20} 80.5 &\cellcolor{gray!20} 81.7 & \cellcolor{gray!20}61.7 & \cellcolor{gray!20}18.8 & \cellcolor{gray!20} 24.6 \\
\midrule
PRPSearcher & \cellcolor{violet!10}24.0 & 26.6 & \cellcolor{violet!10}19.3 & 90.2 & 324.6 \\
UAV-on & 7.4 & 33.3 & 3.8 & \cellcolor{violet!10}64.9 & 364.13 \\
FlySearch & 20.3 & \cellcolor{violet!10}34.2 & 8.6 & 137.5 & 767.7 \\
STARSearcher & 2.4 & 4.9 & 1.1 & 103.5 & \cellcolor{violet!10}296.7 \\
\sysname(Ours) &\textbf{\cellcolor{cyan!15}73.1} & \textbf{\cellcolor{cyan!15}80.5} & \textbf{\cellcolor{cyan!15}42} & \textbf{\cellcolor{cyan!15}11.6} & \textbf{\cellcolor{cyan!15}120.8} \\
\bottomrule
\end{tabular}
}
\label{tab:overall}
\vspace{-18pt}
\end{table}

\subsection{Benchmark Results}\label{VI-B}
\subsubsection{Quantitative Results}
We evaluate \sysname across all scenes as summarized in \tab~\ref{tab:overall}. \sysname demonstrates superior performance across diverse environments over all baseline methods except the Human Agent. Notably, \sysname achieves a 49.1\% improvement in SR compared to the strongest baseline PRPSearcher. 
Among the baselines, STARSearcher exhibits the poorest performance in SR, as it requires close-range inspection of all occupied voxel surfaces. 
While this exhaustive strategy excels in small-scale environments, it makes the drone prone to becoming trapped in structurally complex regions. UAV-on also achieves low SR, as its reliance on salient visual features for decision-making leads to limited active exploration capability.

\begin{figure*}[t]
    \centering
    \setlength{\abovecaptionskip}{-3pt}  
    
    \includegraphics[width=0.98\linewidth]{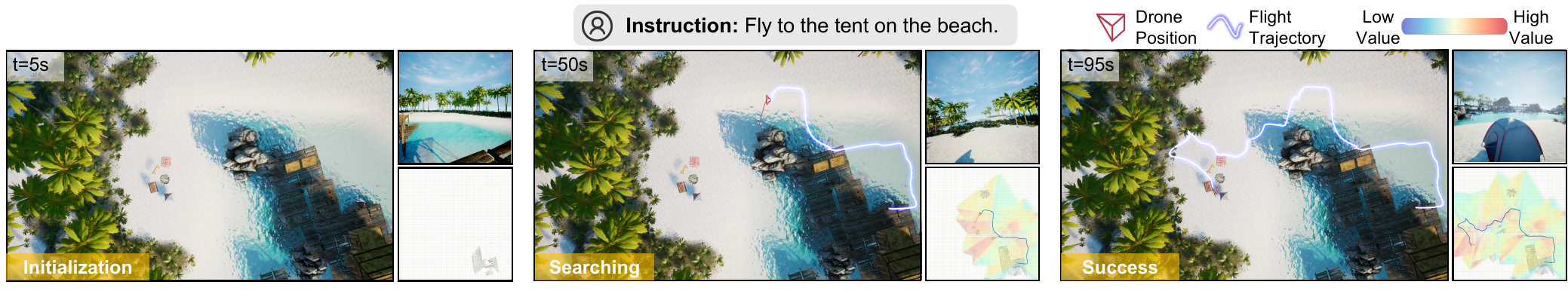}
    \caption{Example qualitative results of \sysname in the Cabinlake environment. The left subfigure shows the top-down view of the scenario. The upper right shows the first-person view of drone and the lower right shows the 3D value map.}
    \label{fig:simdemo}
    \vspace{-12pt}
\end{figure*}
\begin{table}[t]
\caption{Ablation Study on Key Modules.}
\vspace{-5pt}
\centering
\renewcommand{\arraystretch}{1.25}
\resizebox{\columnwidth}{!}{%
\begin{tabular}{lccccc}
\toprule
\textbf{Method}
& \textbf{SR $\uparrow$}
& \textbf{OSR $\uparrow$}
& \textbf{SPL $\uparrow$}
& \textbf{DTG $\downarrow$}
& \textbf{FT $\downarrow$} \\
\midrule
RecTask1              &  {59.3} &  {68.8} &  {40.7} &  {25.6} & 128.7 \\
RecTask2           &  {31.2} &  {38.4} &  {12.5} &  {65.1} & 135.6 \\
\midrule
Value-greedy              &  {16.2} &  {18.3} &  {5.5} &  {105.6} & 312.7 \\
Scalarization              &  {33.2} &  {38.9} &  {17.5} &  {54.2} & 239.4 \\
Two-stage           &  {58.9} &  {63.4} &  {32.5} &  {21.4} & 198.2 \\
\midrule
\rowcolor{gray!20}
\sysname  &  \textbf{73.1} &  \textbf{80.5} &  \textbf{42.0} &  \textbf{11.6} & \textbf{120.8} \\
\bottomrule
\end{tabular}%
}
\label{tab:ablation}
\vspace{-12pt}
\end{table}
Navigation accuracy is further assessed using DTG. \sysname achieves the lowest DTG across all methods.
This advantage stems from the proposed ADTR module, which constructs task-aware keyframes to capture both object-level details and scene-level contextual information, thereby enabling more reliable VLM-based decision-making. 
In contrast, baseline methods fail to consistently integrate both geometric and semantic information into path planning, causing the drone to become disoriented in expansive environments. Note that \sysname surpasses even the Human Agent, as it can obtain the target's 3D position to determine the stopping location more precisely. 
These results demonstrate that \sysname enables accurate target localization in open-world outdoor environments.

We further evaluate navigation efficiency using SPL and FT.
Results indicate that baseline methods demand substantially longer flight times and travel distances to complete tasks. PRPSearcher achieves the highest SR among baselines, yet requires an average FT of 324.6s, 2.5 times longer than \sysname.
This is because PRPSearcher's stop-and-infer behavior, where the drone needs to hover during VLM inference, significantly disrupts flight continuity and reduces efficiency.
In contrast, our asynchronous architecture allows the VLM reasoning and planning processes to operate without blocking each other, achieving more smooth and efficient flight.

\subsubsection{Qualitative Results}
As shown in \fig\ref{fig:simdemo}, we present a representative task episode to illustrate the qualitative results of \sysname, including the executed flight trajectory and 3D value map visualization. Given the instruction ``Fly to the tent on the beach'', the drone starts near a cabin, then generates a smooth trajectory to quickly explore high-value regions, and successfully locates the target object on the beach. The value map shows that \sysname assigns higher values to regions more likely to contain the target object. Moreover, the flight trajectory prioritizes these high-value regions while avoiding revisits to explored areas, demonstrating the spatial consistency of our approach.

\subsection{Ablation Study}
\subsubsection{Ablation Study on ADTR}
In \tab~\ref{tab:ablation}, we evaluate the impact of ADTR on system performance. We design two variants: (i) RecTask1, which replaces coverage-aware keyframe construction with the most recent egocentric observations for Task 1 VLM queries; and (ii) RecTask2, which directly feeds raw observations to the VLM to determine whether the target object is visible and within the success threshold distance. All other components remain identical to \sysname. 
RecTask1 shows significant degradation, with SR dropping from 73.1\% to 59.3\%. This occurs because relying solely on recent observations results in both insufficient spatial coverage and temporal redundancy across consecutive frames, limiting the VLM's spatial reasoning effectiveness.
RecTask2 exhibits even larger degradation, with SR decreasing by 41.9\%. This is expected since VLM cannot reliably estimate the distance to the target object from raw image, leading to premature task termination, particularly in complex scenes with severe occlusion.
These results confirm that our ADTR module is critical for the effectiveness of the downstream planner by providing compact and informative content for VLM reasoning.

\begin{table}[t]
\caption{Average Computation Time ({\normalfont ms}) of Key Components.}
\vspace{-5pt}
\centering
\renewcommand{\arraystretch}{1.2}
\resizebox{\columnwidth}{!}{%
\begin{tabular}{lccccccc}
\toprule
\multirow{3}{*}{\textbf{Scene}} & \multicolumn{3}{c}{\textbf{ADTR \S\ref{IV}}} & \multicolumn{3}{c}{\textbf{SGCP \S\ref{V}}} \\
\cmidrule(lr){2-4}\cmidrule(lr){5-7}

 & \makecell{\textbf{Task 1}\\\textbf{\S \ref{IV-B}}}
 & \makecell{\textbf{Task 2}\\\textbf{\S \ref{IV-B}}}
  & \makecell{\textbf{Value.}\\\textbf{\S \ref{IV-C}}}
 & \makecell{\textbf{Front.+VP.}\\\textbf{\S \ref{V-A},\ref{V-B}}} 
 & \makecell{\textbf{Global}\\\textbf{\S \ref{V-C}}}
 & \makecell{\textbf{Local}\\\textbf{\S \ref{V-D}}}\\
\midrule
Cabinlake
& 3041 & 4711 & 5 & 1.3 & 24.3 & 0.2  \\
Downtown
& 3418 & 4288 & 12 & 2.2 & 13.2 & 0.2  \\
Neighborhood
& 4285 & 3950 & 8 & 2.4 & 30.4 & 0.4 \\
Venice
& 3721 & 5270 & 7 & 3.2 & 14.9 & 0.9 \\
\bottomrule
\end{tabular}%
}
\label{tab:comptime}
\vspace{-10pt}
\end{table}
\subsubsection{Ablation Study on SGCP}
We also evaluate the effectiveness of the SGCP module by comparing three variants: (\textit{i}) Value-greedy, which chooses the frontier with the highest semantic value as the next goal; 
(\textit{ii}) Scalarization, which always selects the frontier with the highest utility score (\ie, value/distance) as the next goal;
and (\textit{iii}) Two-stage~\cite{zhang2025apexnav}, which selects a subset of high-value frontiers by threshold and solves a Traveling Salesperson Problem (TSP) tour over the subset. All other components remain identical to \sysname. 
As shown in \tab~\ref{tab:ablation}, all variants show performance decline across all metrics.
Value-greedy degrades significantly with 160.3\% longer FT and 56.9\% lower SR, since its myopic planning lacks global consideration and thus results in unnecessary revisits. 
Scalarization also underperforms, as it tends to delay distant but substantially higher-value regions, failing to effectively leverage the semantic guidance.
Two-stage performs best among the baselines but still shows a substantial degradation, with SR dropping to 58.9\% and FT increasing to 198.2s. 
This is because its hard frontier selection threshold assumes high-value regions necessarily contain the target, potentially excluding regions where the target actually resides. Additionally, TSP's distance-minimization objective may defer exploring high-value but distant regions.
In contrast, our algorithm selectively establishes precedence constraints that ensure frontiers with significantly higher values are prioritized regardless of distance, while geometric proximity determines ordering among similarly-valued frontiers, enabling a more flexible balance between semantic prioritization and geometric travel efficiency.

\subsection{Computation Time }
We evaluate the computation time of key components, measured per update for ADTR and per planning iteration for SGCP.
Table~\ref{tab:comptime} shows that our hierarchical SGCP module achieves substantial computational efficiency. Even the global planning component (\S\ref{IV-C}), which dominates the computation budget, requires less than 31ms per iteration. 
\sysname exhibits higher SGCP computation time in complex environments like Neighborhood, since dense tree obstacles increase path planning complexity.
These results demonstrate that our algorithm provides effective solutions while maintaining computational efficiency.
Notably, a single VLM query in ADTR can take two orders of magnitude longer than the entire planning cycle. This highlights the importance of decoupling continuous planning from VLM reasoning. Without this asynchronous architecture, high-latency VLM queries would bottleneck drone agility and cause mission failures. 

\begin{figure}[t]
    \centering
    \setlength{\abovecaptionskip}{-10pt}  

    \includegraphics[width=0.98\linewidth]{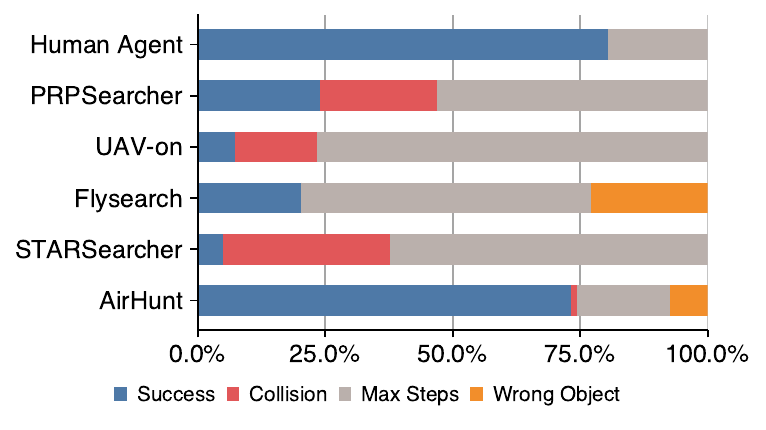}
    \caption{Failure Breakdown of \sysname and Baselines.}
    \label{fig:breakdown}
    \vspace{-15pt}
\end{figure}
\begin{figure}[t]
    \centering
    \setlength{\abovecaptionskip}{-8pt}  
    \includegraphics[width=0.98\linewidth]{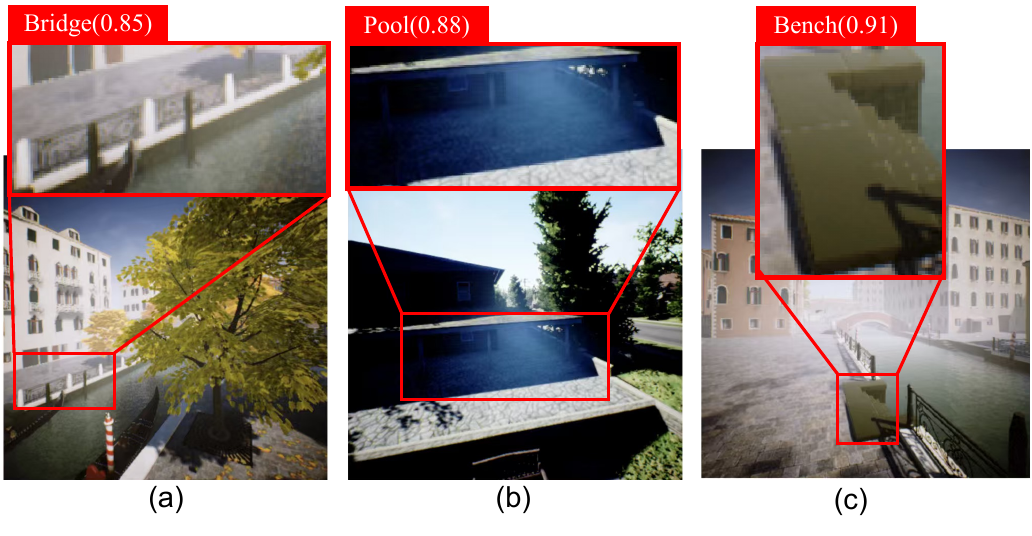}
    \caption{Examples of misclassification by object detectors. (a) A fence is misclassified as a bridge due to texture similarity. (b) A shadowed area is incorrectly detected as a pool. (c) A staircase is misidentified as a bench.}
    \label{fig:failedcase}
    \vspace{-8pt}
\end{figure}
\subsection{Failure Breakdown}
We analyze and compare the failure reasons of all methods in the experiments, as illustrated in \fig\ref{fig:breakdown}. The categories include: \textit{Success}, indicating task completion; \textit{Collision}, referring to instances where the drone triggers a collision with an obstacle; \textit{Max Steps}, where the drone exceeds the maximum number of VLM queries without locating the target object; and \textit{Wrong Object}, where the drone stops in front of an incorrect object. As shown, the Human Agent achieves the best performance, closely followed by \sysname. Three key observations emerge from the analysis: 
(i) Baseline methods primarily fail due to exceeding the maximum steps. This is attributed to their over-reliance on VLM for decision-making, which often leads to inefficient exploration, as discussed in \S\ref{VI-B}; 
(ii) Most baselines exhibit considerable collision rates that would be unacceptable in real-world deployments where safety is critical. In particular, STARSearcher inspects every occupied surface at close distances, leading to frequent collisions with complex outdoor obstacles such as trees and thin lampposts. Other methods encounter collisions due to discrete motion primitives and inability to replan upon receiving new sensor observations. In contrast, \sysname substantially reduces collisions through elaborate planning, though it occasionally collides when exploring extremely narrow regions such as dense forests; 
(iii) Although our system is highly effective, it inherits certain limitations from object detector, Grounding-DINO~\cite{liu2024grounding}. As shown in \fig\ref{fig:failedcase}, the detector may hallucinate objects or misclassify background elements (e.g., shadows or fences) with high confidence due to texture ambiguity or lighting variations. This aligns with our real-world experiments, confirming that the primary bottleneck lies in the detector's sensitivity to visual noise.

\begin{figure*}[t]
    \centering
    \setlength{\abovecaptionskip}{-3pt}  
    
    \includegraphics[width=0.98\linewidth]{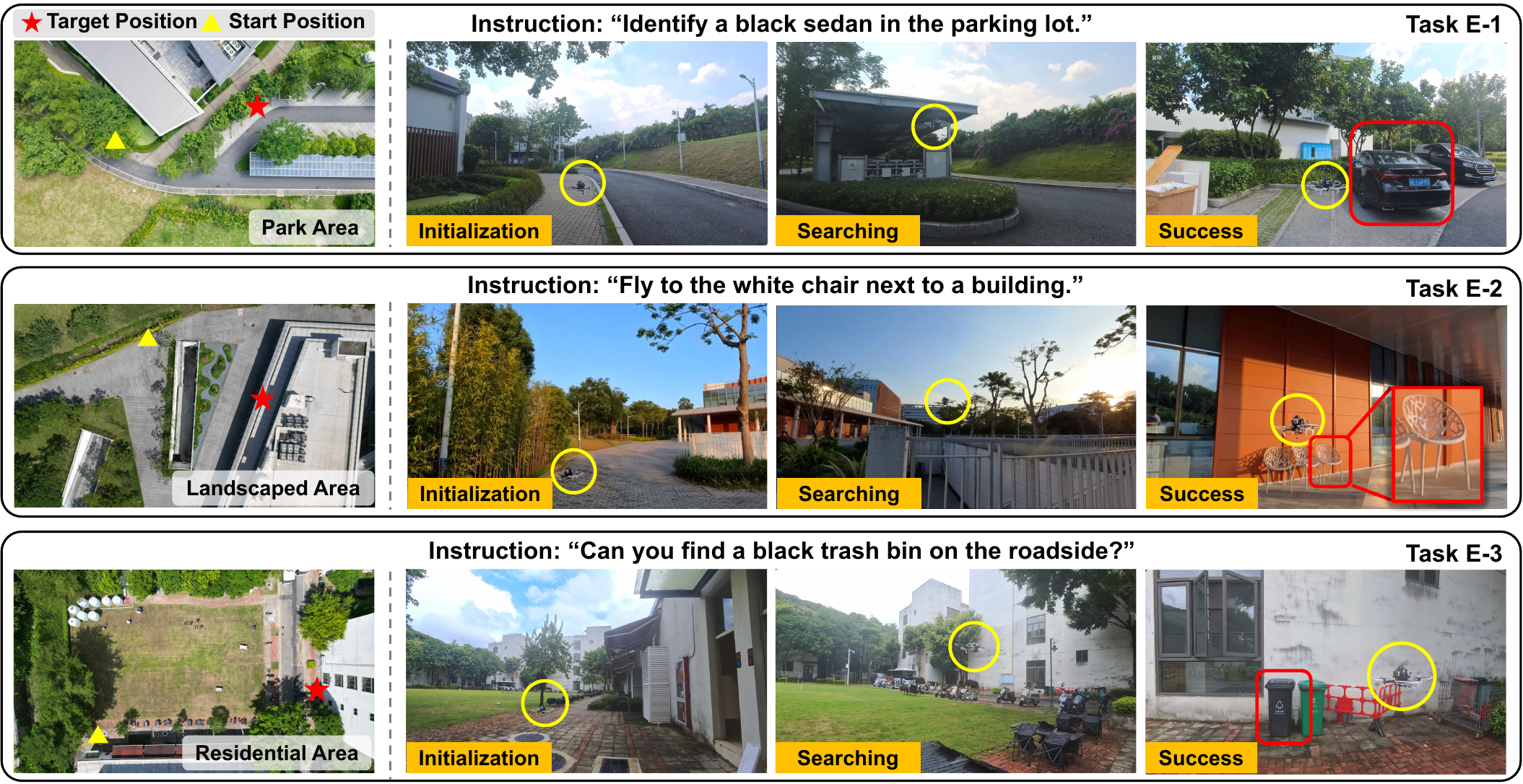}
    \caption{Real-world experiments are conducted in three challenging environments, including a Park Area (E-1), a Landscaped Area (E-2), and a Residential Area (E-3). For each task, the first image illustrates a top-down view of the experimental scenario, and the others are snapshots of different drone states during task execution. The yellow circle marks the drone, and the red rectangles mark the target objects.}
    \label{fig:realworld}
    \vspace{-10pt}
\end{figure*}
\begin{figure}[t]
    \centering
    \begin{minipage}[t]{0.40\linewidth}
        \setlength{\abovecaptionskip}{-10pt}  

        \centering
        \includegraphics[width=0.98\linewidth]{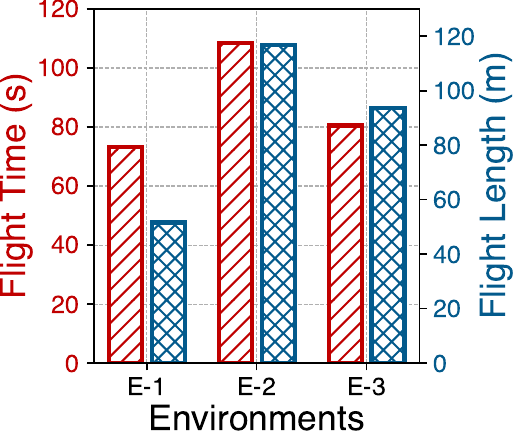}
        \caption{Total flight time and flight length in three real-world experimental tasks.}
        \label{fig:results_rw}
    \end{minipage}
    \hfill
    \begin{minipage}[t]{0.58\linewidth}
        \centering
        \setlength{\abovecaptionskip}{-10pt}  

        \includegraphics[width=0.98\linewidth]{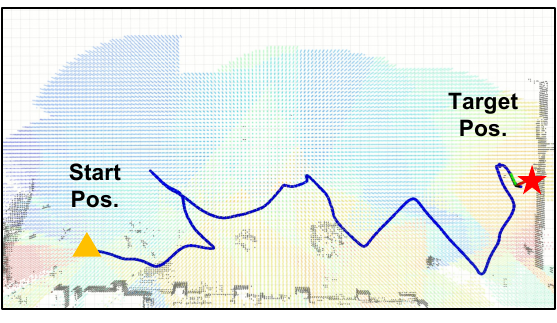}
        \caption{Example of drone's executed trajectory and value map in real-world experiments.}
        \label{fig:traj}
    \end{minipage}
    \vspace{-15pt}
\end{figure}
\section{Real-world Experiments}
\subsubsection{Experimental Setup}
We conduct field experiments using a customized quadrotor platform equipped with an Intel NUC 11 Pro (8GB), a NxtPX4v2 autopilot, a Mid360 LiDAR, and three OAK-4P-New cameras with forward, left, and right orientations. The system operates under dynamics constraints of $v_{m}=1.0$m/s, $a_m=1.0$m/s$^2$, and $\omega=1.05$rad/s. We employ FAST-LIO~\cite{xu2021fast} for onboard localization and mapping. A ground station laptop communicates with the quadrotor and queries the online VLM via the 4G cellular network.

\begin{figure}[t]
    \centering
    \setlength{\abovecaptionskip}{-5pt}  
    \includegraphics[width=0.98\linewidth]{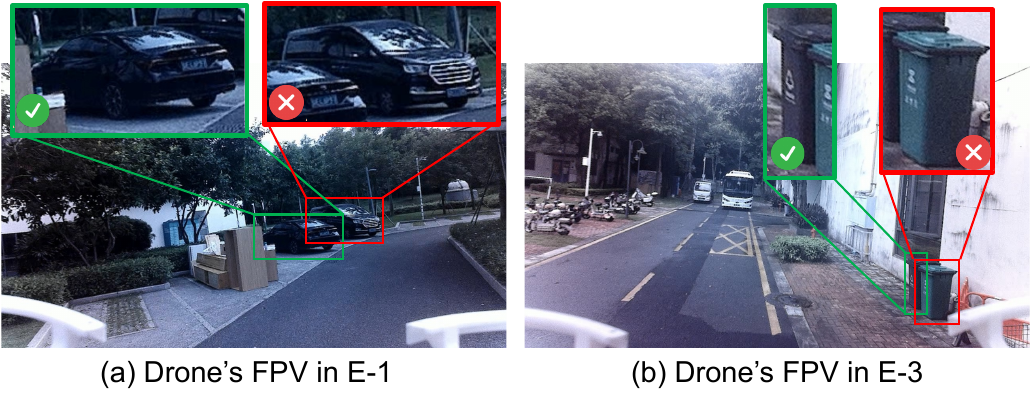}
    \caption{Snapshots of the drone's FPV when detecting relevant objects in (a) E-1 and (b) E-3. (a) \sysname successfully identifies the target sedan (green box) and excludes the MPV car (red box), which is similar but incorrect. (b) \sysname correctly identifies the target black bin (green box) while excluding the green bin (red box).}
    \label{fig:fpv}
    \vspace{-10pt}
\end{figure}

We evaluate \sysname on three tasks in challenging outdoor environments with open-set instructions, as illustrated in Fig.\ref{fig:realworld}. These scenarios exhibit variations in environmental layout and navigational context, with distinct spatial characteristics including obstacle density, open space ratio, and terrain elevation. Specifically,
\textbf{Task E-1} is conducted in a park area with residential buildings, tree-lined curved streets, and parking spaces. The drone navigates from a complex intersection to locate a black sedan.
\textbf{Task E-2} is performed in a landscaped area with obstacles including walls and barriers. The drone starts from a position adjacent to a barrier to locate a white chair.
\textbf{Task E-3} takes place in a residential zone with a spacious grassy field surrounded by roads. The drone starts from a corner of the grassland to locate a black trash bin.

\subsubsection{Quantitative Results}
\fig\ref{fig:results_rw} summarizes the quantitative performance of \sysname across all experimental scenarios. The system successfully completes navigation tasks in tasks E-1, E-2, and E-3 with flight times of 73.2s, 108.5s, and 80.5s, respectively. The example trajectory illustrated in \fig\ref{fig:traj} demonstrates continuous execution with minimal detours and limited revisits to previously explored regions, resulting in efficient navigation with short flight durations and distances. This efficiency can be attributed to our SGCP module, which fuses geometric information and semantic cues to generate consistent trajectories.

\subsubsection{Qualitative Results}
We present the experimental scenes and the navigation process from initialization through searching to task completion, demonstrating successful target localization in all scenarios based on given instructions (\fig\ref{fig:realworld}). \fig\ref{fig:fpv} further illustrates snapshots of the drone's first-person view when detecting objects relevant to the given instructions. \sysname's ADTR module incorporates a robust object verification mechanism that successfully excludes visually similar but semantically incorrect objects and identifies the correct target that matches the semantic context in the instructions. This verification capability contributes to increased navigation success rate in complex real-world environments. In summary, these results validate \sysname's superior object navigation capabilities in large-scale outdoor environments. Detailed experimental results and additional visualizations are provided in the Supplementary Material.

\section{Conclusion}
This paper introduces \sysname, an aerial object navigation system that navigates smoothly and efficiently to locate open-set objects based on open-set instructions, with zero-shot generalization to various outdoor scenarios.
At the core of \sysname is the asynchronous system architecture, which enables the drone to maintain continuous flight while incorporating intermittent guidance from the VLM. 
We further design two modules, \textit{Active Dual-Task Reasoning} and \textit{Semantic-Geometric Coherent Planning}, that work synergistically to enhance search efficiency and consistency.
Extensive evaluations in diverse simulation environments and complex real-world scenarios demonstrate its superior performance.

\bibliographystyle{IEEEtran}
\bibliography{ref}


\newpage

 




\vfill

\end{document}